\documentclass[10pt,conference]{IEEEtran}

\usepackage{graphicx}
\usepackage{amsmath,amssymb}
\usepackage{cite}
\usepackage{hyperref}
\hypersetup{
    colorlinks=true,
    linkcolor=blue,
    citecolor=blue,
    urlcolor=blue
}

\usepackage[margin=1in]{geometry}

\usepackage{graphicx}
\usepackage{tikz}
\usepackage{cleveref}
\usepackage{booktabs}
\newcommand{\eg}{e.g.,\ }

\title{SAM-PTx: Text-Guided Fine-Tuning of SAM with Parameter-Efficient, Parallel-Text Adapters}

\author{
  \IEEEauthorblockN{Shayan Jalilian}
  \IEEEauthorblockA{
    University of Regina\\
    \texttt{sjs949@uregina.ca}
  }
  \and
  \IEEEauthorblockN{Abdul Bais}
  \IEEEauthorblockA{
    University of Regina\\
    \texttt{abdul.bais@uregina.ca}
  }
}

\date{}

\begin{document}
\maketitle
\begin{abstract}
The Segment Anything Model (SAM) has demonstrated impressive generalization in prompt-based segmentation. Yet, the potential of semantic text prompts remains underexplored compared to traditional spatial prompts like points and boxes. This paper introduces SAM-PTx, a parameter-efficient approach for adapting SAM using frozen CLIP-derived text embeddings as class-level semantic guidance. Specifically, we propose a lightweight adapter design called Parallel-Text that injects text embeddings into SAM’s image encoder, enabling semantics-guided segmentation while keeping most of the original architecture frozen. Our adapter modifies only the MLP-parallel branch of each transformer block, preserving the attention pathway for spatial reasoning. Through supervised experiments and ablations on the COD10K dataset as well as low-data subsets of COCO and ADE20K, we show that incorporating fixed text embeddings as input improves segmentation performance over purely spatial prompt baselines. To our knowledge, this is the first work to use text prompts for segmentation on the COD10K dataset. These results suggest that integrating semantic conditioning into SAM’s architecture offers a practical and scalable path for efficient adaptation with minimal computational complexity.
\end{abstract}

\section{Introduction}
\label{sec:intro}

Semantic segmentation requires dense pixel-level supervision, making it a costly and time-intensive process. As a result, recent research has focused on leveraging foundation models to reduce labelling burden and improve generalization across visual domains. The Segment Anything Model (SAM)~\cite{kirillov2023segment} has demonstrated remarkable zero-shot segmentation capabilities through its prompt-based design, accepting spatial prompts such as points, boxes, or masks to guide predictions.

While SAM’s spatial prompt mechanism is highly effective for directing segmentation to specific regions, it operates without access to high-level semantic information. In practice, selecting a meaningful spatial prompt often assumes prior knowledge of the target class; for example, placing a point on a "person" requires the user to recognize people in the image visually. Additionally, spatial prompts typically guide segmentation toward a single instance or region, making capturing all objects of the same semantic category difficult without carefully constructed prompting strategies. By introducing text embeddings as semantic cues while fine-tuning SAM, our method enhances spatial prompting with global semantics-level guidance, enabling more flexible and semantically aware segmentation.

This work investigates the potential of combining semantic text guidance with traditional spatial prompts to enhance segmentation. Specifically, although we use point-based spatial prompts during training, we show that augmenting SAM with frozen CLIP~\cite{radford2021learning} text embeddings as additional guidance improves its ability to perform class-aware segmentation. We name our method \textbf{SAM-PTx}, short for \emph{SAM with Parallel-Text Adapters}, where the text-conditioned adapters are inserted in parallel to the MLP pathway, following prior PEFT designs~\cite{song2024SUSAM}.

\begin{figure}[t]
  \centering
  \fbox{\includegraphics[width=\linewidth]{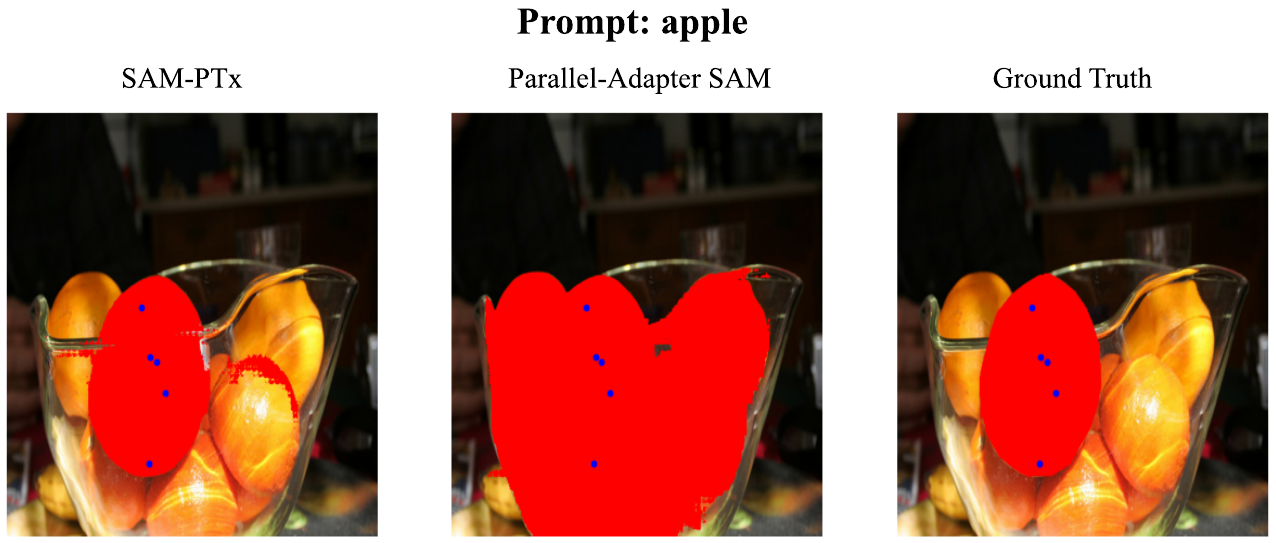}}
  \caption{Qualitative comparison for the class “apple” in a visually ambiguous scene containing multiple fruits. Both models receive identical point prompts (blue), but only our method correctly segments the apples (red) by leveraging semantic input from the text prompt. In contrast, the non-text-guided SAM segments all objects, underscoring the value of incorporating semantic guidance.}
  \label{fig:intro_compare}
\end{figure}

Vision-language models like CLIP~\cite{radford2021learning} align images and text in a shared semantic space and have shown strong performance in tasks like zero-shot classification, and weakly supervised segmentation~\cite{Lin2023CLIPEffSeg, Zhou2023ZegClip}. Yet, despite SAM’s flexibility and the potential of CLIP embeddings as semantic prompts, existing adaptations of SAM rarely explore the integration of textual guidance during training, especially in a modular and parameter-efficient way.

In this work, we propose a lightweight, parameter-efficient adapter design that enables semantically guided fine-tuning of SAM by injecting frozen CLIP-derived text embeddings into its image encoder. Unlike prior PEFT methods that rely solely on spatial prompts~\cite{chen2023SamAdptr,song2024SUSAM}, our approach incorporates class-level semantic conditioning, allowing the model to leverage both spatial and textual cues during training. The majority of SAM’s architecture remains frozen, and text embeddings are precomputed per class, making our method both modular and efficient.

We conduct supervised experiments on subsets of standard benchmarks such as ADE20K and COCO, chosen to simulate extreme low-data settings and to reduce computational cost. Despite using limited data, SAM-PTx outperforms non-text-enhanced fine-tuning and improves segmentation quality. Our method represents a step toward making SAM adaptable to semantic prompts, laying the foundation for future multimodal extensions.
\section{Related Work}
\label{sec:related}

\subsection{Prompt-based segmentation and SAM}

SAM~\cite{kirillov2023segment} is a powerful foundation model trained on over a billion masks to perform prompt-driven segmentation across diverse visual domains. Its architecture decouples image and prompt encoders, enabling it to process spatial prompts such as points, boxes, or masks to segment specific regions of interest. While SAM demonstrates strong generalization and zero-shot capabilities, its performance can still be limited by the precision and informativeness of the input prompts. For downstream or domain-specific tasks, further adaptation, such as prompt optimization or fine-tuning, is often required to meet task-specific requirements.

Several works have investigated adapting SAM to new domains ~\cite{chen2023SamAdptr, zhong2024convMLoRA, song2024SUSAM}. While these methods improve performance through parameter-efficient fine-tuning techniques, they rely exclusively on spatial prompts and do not incorporate any form of semantic or text-based guidance. As a result, the model’s understanding remains grounded in spatial localization cues, without leveraging class-level semantics or language supervision to enhance segmentation.

In contrast, our work aims not to eliminate spatial prompts but to augment them with semantic information through CLIP-derived text embeddings. We show that injecting such information into SAM’s image encoder enhances its ability to learn class-aware segmentation behaviour, improving its robustness and adaptability in supervised fine-tuning settings.

\subsection{Parameter-efficient fine-tuning (PEFT)}

To reduce the cost of adapting large models to downstream tasks, parameter-efficient fine-tuning (PEFT) techniques have become increasingly popular. Instead of updating all model weights, PEFT methods introduce lightweight, trainable components---such as adapters or low-rank updates---that enable effective fine-tuning with significantly fewer parameters. Representative approaches include LoRA~\cite{hu2022lora}, Visual Prompt Tuning (VPT)~\cite{jia2022vpt}, and adapter-based tuning~\cite{houlsby2019parameter}.

In the context of SAM, adapter-based PEFT was first introduced by~\cite{chen2023SamAdptr}, who incorporated lightweight adapter modules into SAM’s architecture to enable efficient domain-specific adaptation. While their method demonstrated strong performance on targeted tasks, it required manually designed input features. It lacked generalizability across domains, partly due to its reliance on post-processing and more complex and manually-crafted prompting pipelines.

Subsequently, SU-SAM~\cite{song2024SUSAM} introduced a simple and unified PEFT framework for SAM, systematically evaluating different combinations of adapter and LoRA configurations. The mixed adapter showed the best performance on their benchmarks among the four variants they proposed. However, our preliminary evaluations found that the parallel adapter—a simpler and more modular variant—consistently outperformed the others. This motivated us to adopt the parallel adapter as the foundation for our method.

One of SU-SAM’s key strengths is its balance of simplicity, effectiveness, and generalizability. Compared to earlier methods like SAM-Adapter~\cite{chen2023SamAdptr}, which relied on hand-crafted input features and complex prompting pipelines, or Conv-Meets-LoRA~\cite{zhong2024convMLoRA}, which introduced convolutional refinements with mixture-of-experts for improved efficiency in image-centric tasks, SU-SAM offers a more lightweight and broadly applicable solution. Its minimal architectural modifications make it especially well-suited for our goal of injecting semantic information into SAM in a parameter-efficient and modular way.

While a few recent works have explored incorporating text information into SAM, primarily by feeding embeddings into prompt encoders or using them to filter SAM outputs, integrating CLIP-derived text embeddings directly into SAM’s image encoder in a modular, parameter-efficient manner remains largely unexplored.

Building on the parallel adapter structure from SU-SAM, our method is the first to inject CLIP-derived text embeddings into SAM’s image encoder for semantic conditioning. Unlike prior approaches that rely on prompt engineering or post-hoc filtering, our design enables joint training of spatial prompts and semantic embeddings in a streamlined setup. This structure preserves most of SAM’s original architecture while offering a lightweight and effective strategy for incorporating vision-language alignment during fine-tuning.

\subsection{Integrating text inputs with SAM}

Recent efforts have explored augmenting SAM with textual information, primarily through integrations with vision-language models (VLMs) like CLIP or GroundingDINO~\cite{liu2023groundingDINO}. In these works, CLIP is mostly used to generate or embed textual prompts that can be passed to SAM in various forms, or as a guide that filters the outputs of SAM. These strategies focus on how to incorporate textual cues into SAM’s segmentation pipeline, and can be grouped into three primary categories:

\textbf{Text-to-spatial prompt generation:} These approaches use a VLM to transform text prompts into spatial cues (\eg, boxes or points) that are then passed to SAM. This strategy is adopted by Grounded-SAM~\cite{ren2024groundedSAM}, MedCLIP-SAMv2~\cite{koleilat2409medclipsamv2}, CLIPSAM~\cite{li2025clipsam}, SAM2CLIP2SAM~\cite{kollias2024sam2clip2sam}, CLIP-Guided SAM Adaptation~\cite{chen2024clipguidedsam}, and CLIPSurgery~\cite{li2025CLIPSurgery}, which generates point prompts from CLIP similarity maps and feeds them to SAM.

\textbf{CLIP-guided mask selection from SAM outputs:} These methods run SAM in everything mode to generate a set of candidate masks, then rank or filter them based on similarity between CLIP-derived text features and mask/image features. This approach is used in Semantic Segment Anything (SSA)~\cite{chen2023semanticSAM}, Segment Anything with CLIP~\cite{curtpark2023SAMwCLIP}, SaLIP~\cite{aleem2024salip}, and SAM as the Guide~\cite{yang2024samasguide}.

\textbf{Text embeddings as sparse prompts for the prompt encoder:} These works encode class names or referring expressions using CLIP’s text encoder and feed the resulting embeddings into SAM’s prompt encoder or a modified or custom-designed prompt encoder as sparse prompts. Examples include RefSAM~\cite{li2023refsam}, EVF-SAM~\cite{zhang2024evfsam}, AdaptiveSAM~\cite{paranjape2024adaptivesam}, S-SAM~\cite{paranjape2024ssam}, and the original SAM paper~\cite{kirillov2023segment}, which discusses this capability but does not release full support in the official code.

While these strategies highlight the versatility of combining SAM with vision-language supervision, they differ fundamentally from our approach in intent and integration point.

First, methods that convert text into spatial prompts (\eg, via CLIP or GroundingDINO) are orthogonal to ours: they focus on prompt automation and typically do not fine-tune SAM, whereas our method focuses on improving SAM's performance through fine-tuning by injecting semantic text information into the image encoder. Our method can coexist with such spatial-prompt generators, as we remain agnostic to the source of spatial inputs.

Second, post-hoc filtering approaches rely on SAM’s everything mode to produce candidate masks, which are later ranked or filtered using CLIP. These techniques operate independently from SAM’s internal representations and generally avoid fine-tuning; in contrast, we directly modify SAM’s representation learning by injecting text embeddings during training to enhance segmentation quality.

Third, prior works that input text embeddings into SAM’s prompt encoder (\eg, RefSAM, EVF-SAM, AdaptiveSAM) are closer in spirit to ours but differ in design and objective. These works treat text as an external prompt, often without modifying SAM’s image encoder or decoder. In contrast, we inject frozen CLIP-derived embeddings into SAM’s image encoder and fine-tune the model to align visual and textual features better internally. This enables richer multimodal conditioning while preserving SAM’s modular structure.

To our knowledge, this is the first method to demonstrate that frozen text embeddings, when injected directly into SAM’s image encoder, can lead to measurable segmentation improvements under supervised fine-tuning, all while requiring minimal architectural changes.

\section{Method}
\label{sec:method}

We refer to our approach as SAM-PTx, which augments SAM by injecting \emph{frozen} CLIP-derived text embeddings into its image encoder using a simple yet effective design we call the \textit{Parallel-Text Adapter}. While inspired by prior parallel adapter techniques~\cite{song2024SUSAM}, our adapter is specifically tailored to introduce semantic understanding into SAM’s visual pipeline with minimal structural changes and without adding many new parameters. SAM-PTx preserves the standard point-based prompt mechanism and enhances segmentation quality by enriching the model’s internal representations with class-level text guidance. It achieves this while training only a small fraction of the total parameters and introducing a new modality in the form of text prompts.

\subsection{Overview}
\label{sec:overview}

\begin{figure*}[t]
  \centering
  \includegraphics[width=0.9\linewidth]{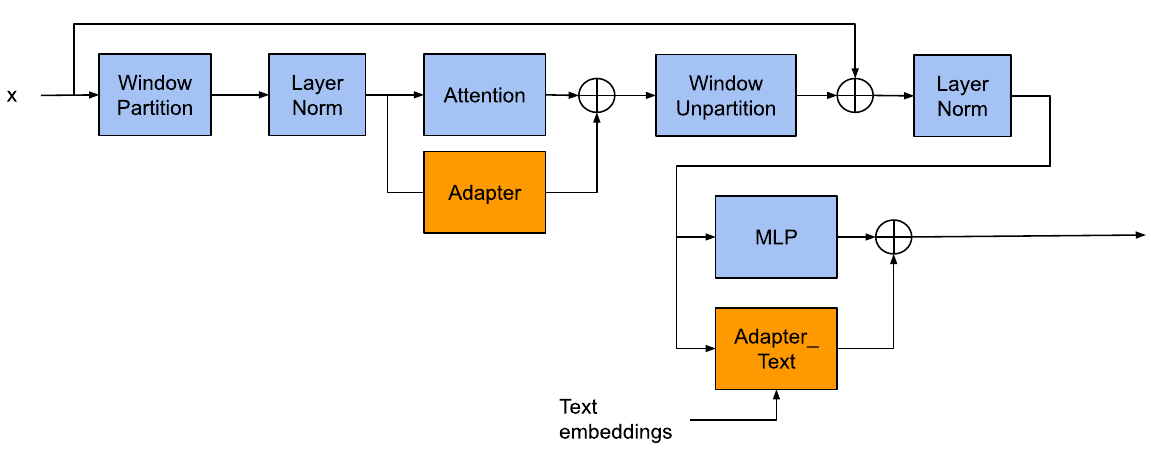}
  \caption{Architecture overview.  CLIP text embedding (blue) is injected into every transformer block via Parallel‑Text Adapters (orange). Only the orange blocks are trainable.}
  \label{fig:arch_overview}
\end{figure*}

Given an image~$I$, a spatial prompt~$p$ (e.g., a foreground point), and a class label~$\ell$, we compute the CLIP text embedding $t = f_{\text{text}}(\ell)$ and inject it into the SAM’s image encoder through lightweight adapters (\cref{fig:arch_overview}). During supervised fine-tuning, only the adapter weights and the mask decoder are updated; all other SAM weights remain frozen. The text embeddings are precomputed and cached as part of the training data.

At inference time, the same mechanism produces segmentation masks that reflect \emph{both} spatial and semantic cues.

\subsection{Parallel-Text Adapter}
\label{sec:adapter}

\begin{figure}[t]
  \centering
  \includegraphics[width=0.9\linewidth]{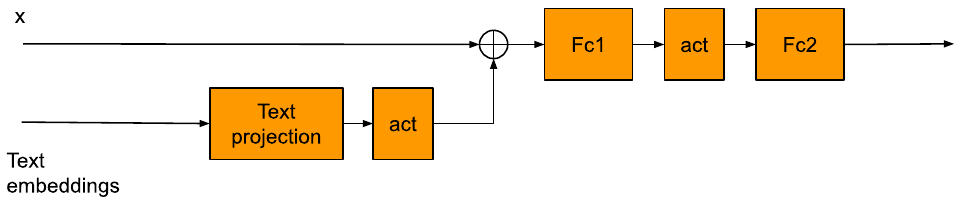}
  \caption{Our Parallel-Text Adapter. The text projection, followed by the non-linear activation, is added to the input and goes through the Parallel Adapter from SU-SAM~\cite{song2024SUSAM}.}
  \label{fig:block_diagram}
\end{figure}

We build on the \emph{parallel adapter} design introduced in SU-SAM~\cite{song2024SUSAM}, which places lightweight adapters in parallel with both the multi-head self-attention (MHSA) and MLP blocks in each transformer layer of SAM’s ViT-based image encoder (see \cref{fig:block_diagram}).

Our method retains this dual-branch structure: the MHSA-parallel and MLP-parallel adapters are trained. However, only the adapter parallel to the MLP block is extended to incorporate semantic information from text embeddings. The MHSA-parallel adapter remains a standard adapter without text conditioning.

We adopt this selective design for two reasons. First, MLP layers process tokens independently and are well-suited for injecting global semantic context such as class-level text embeddings. Second, the self-attention mechanism is thought to play a critical role in propagating spatial prompts across the image. Modifying this pathway with text could interfere with the core prompt propagation mechanism, reducing effectiveness. We preserve this spatial reasoning by isolating text conditioning to the MLP pathway while introducing high-level semantic alignment.

The standard parallel adapter is defined as:
\begin{equation}
  \text{Adapter}(x) = x + W_{\text{up}} \cdot \sigma\left(W_{\text{down}} x \right),
  \label{eq:vanilla_adapter}
\end{equation}
where $\sigma$ denotes a GELU activation.

\vspace{0.2em}
\noindent\textbf{Semantic extension.}\;
To inject semantic information, we use CLIP-derived text embeddings $t \in \mathbb{R}^{d_t}$, which are first projected into the visual token dimension:
\[
\tilde{t} = \mathrm{Act}(W_t t),
\]
and added to the visual input:
\[
x' = x + \tilde{t}.
\]
This modified input is then passed through a bottleneck MLP:
\begin{equation}
  \text{TextAdapter}(x, t) = W_2 \cdot \mathrm{Act}(W_1 x').
  \label{eq:text_adapter}
\end{equation}

The weights $W_1$, $W_2$, and $W_t$ in the text-conditioned adapters are trainable. Both adapter types (MHSA-parallel and MLP-parallel) are optimized during fine-tuning, but only the latter incorporates semantic guidance from the text embeddings.

This separation allows the model to benefit from text-based conditioning without disrupting spatial reasoning, balancing modularity, parameter efficiency, and prompt interpretability.

\subsection{Semantic Text Embeddings}
\label{sec:text_embed}

We use CLIP ViT‑B/32 and the class name as the prompt
to obtain a 512‑D embedding per class.
Embeddings are cached and reused across images, adding negligible overhead.

\subsection{Training Objective}
\label{sec:loss}

For an image–prompt–label triple $(I,p,\ell)$ with ground‑truth mask
$M_{\text{gt}}$, we predict
\mbox{$M = \mathrm{SAM}(I,p;\,t)$},
where $t$ is the cached embedding of~$\ell$.
We minimize binary cross‑entropy:
\begin{equation}
  \mathcal{L}_{\text{seg}} =
  \mathrm{BCE}\!\bigl(M,\;M_{\text{gt}}\bigr).
  \label{eq:loss}
\end{equation}

\subsection{Inference}
\label{sec:inference}

At test time, the user supplies a spatial prompt and class label.
The text embedding conditions the MLP-parallel adapters via~\eqref{eq:text_adapter},
and the decoder outputs a mask that reflects the query's spatial location and semantic identity.

This design preserves SAM’s prompt‑driven workflow while introducing semantics‑guided segmentation with minimal parameter cost.

\section{Experiments}
\label{sec:experiments}

We evaluate SAM-PTx on two standard segmentation benchmarks — COCO~\cite{lin2014microsoft} and ADE20K~\cite{zhou2017scene} — using low-data subsets that simulate realistic transfer learning scenarios. We aim to demonstrate the benefit of injecting frozen CLIP text embeddings into SAM’s image encoder via adapters, and to assess how semantic guidance improves segmentation performance compared to prompt-only baselines.

\subsection{Datasets}

We evaluate SAM-PTx on two challenging and widely used segmentation benchmarks: COCO and ADE20K. To simulate realistic low-label settings commonly encountered in transfer learning and parameter-efficient fine-tuning, we use small labelled subsets from each dataset, following existing protocols. These subsets reflect practical conditions with limited annotations, as commonly encountered in real-world deployment scenarios, and allow us to assess the effectiveness of SAM-PTx in low-data regimes.

\textbf{COCO}~\cite{lin2014microsoft} is a large-scale object detection, segmentation, and captioning dataset with over 118,000 training images and 80 object categories. For our experiments, we use the \texttt{1\_512} labelled subset from PseudoSeg~\cite{zou2020pseudoseg}, which consists of 232 labelled images covering 20 object categories. Since we formulate the task as binary segmentation per object, each object instance becomes a separate training sample.s This results in a total of 631 binary segmentation training samples for COCO.

\textbf{ADE20K}~\cite{zhou2017scene} is a densely annotated scene parsing dataset containing over 20,000 images and 150 semantic classes. We adopt the \texttt{1\_64} labelled split from SemiVL~\cite{hoyer2024semivl}, which includes 316 labelled images. Like COCO, we treat each object instance as a binary segmentation sample, yielding 2,535 training samples for ADE20K.

These small subsets are used in prior semi-supervised learning works and represent practical fine-tuning conditions where labelled data is limited due to cost or computational constraints. Our setup aligns with these real-world transfer learning scenarios, making the evaluation more relevant and challenging.

\subsection{Experimental Setup}

We evaluate under a supervised fine-tuning setting with spatial point prompts and class-level text labels. We compare the following variants:

\begin{itemize}
    \item \textbf{Vanilla SAM}: The original Segment Anything Model, frozen during training, using only spatial prompts.
    \item \textbf{SU-SAM}~\cite{song2024SUSAM}: A parameter-efficient version of SAM with parallel adapters inserted into the transformer blocks, but without text guidance.
    \item \textbf{SAM-PTx}: SAM-PTx builds on SU-SAM but injects frozen CLIP text embeddings into the image encoder via modified adapters.
\end{itemize}

We focus our comparisons on SU-SAM’s parallel adapter variant, which we selected after running all four variants on our dataset and observing that the parallel version consistently performed best. This variant also serves as the foundation for SAM-PTx's design. We did not include additional prior methods in our evaluation, as current SAM+text approaches rely on prompt generation or post-processing rather than end-to-end fine-tuning with semantic conditioning. As such, our experiments primarily focus on our method's ablations and the impact of semantic text guidance within this unique setup.

\subsection{Implementation Details}

We use the ViT-B version of SAM and freeze the image encoder except for the inserted adapters. We use the CLIP ViT-B/32 text encoder to extract text embeddings with the prompt format “a photo of a \{class\},” following standard CLIP practice.

All models are trained for 30 epochs using the Adam optimizer with a learning rate of $1\mathrm{e}{-5}$ and a batch size of 1. Input images are resized to $1024 \times 1024$, the accepted resolution for SAM. During training and inference, we use five foreground point prompts for each object along with its class name.

To preserve fine-grained object details, we upsample SAM’s default $256 \times 256$ output masks to $512 \times 512$ using bilinear interpolation before computing the loss. This helps avoid label loss due to resizing, particularly for small or thin objects.
All experiments were conducted on two NVIDIA RTX 4090 GPUs.

\subsection{Results}
We evaluate SAM-PTx on the COCO \texttt{1\_512} and ADE20K \texttt{1\_64} subsets, comparing against several baselines, including unmodified SAM, decoder-only fine-tuning, and the standard Parallel adapter from SU-SAM~\cite{song2024SUSAM}.

As shown in Table~\ref{tab:quant_results}, fine-tuning only the SAM decoder already yields a strong performance boost compared to zero-shot SAM. However, our Parallel-Text Adapter design further improves results by incorporating semantic guidance through frozen CLIP embeddings. Despite the simplicity of the adapter design and the low-data setting, SAM-PTx consistently outperforms the purely spatial Parallel-Adapter baseline.

\begin{table}[htbp]
\centering
\begin{tabular}{lcc}
\toprule
Method & COCO \texttt{1\_512} & ADE20K \texttt{1\_64} \\
\midrule
No fine-tuning & 62.09 & 65.14 \\
Decoder-only & 67.29 & 70.32 \\
Parallel & 67.35 & 71.29 \\
\textbf{Parallel-Text} & \textbf{67.77} & \textbf{71.38} \\
\bottomrule
\end{tabular}
\vspace{1mm}
\caption{Segmentation performance (mIoU) on COCO and ADE20K low-data splits. SAM-PTx consistently improves by injecting frozen text embeddings into SAM’s image encoder.}
\label{tab:quant_results}
\end{table}

\subsubsection{Where to Inject Text Embeddings?}

We conducted a design study to explore which component of SAM’s architecture benefits most from semantic guidance. Specifically, we experimented with injecting frozen CLIP text embeddings into the image encoder, the prompt encoder, and the mask decoder, one at a time. In all three cases, the embeddings were projected using a lightweight linear layer followed by a non-linear activation, i.e., $\mathrm{Act}(W_t t)$, where $t$ is the CLIP embedding. Each setup used the same text projection mechanism for consistency.

For the \textbf{prompt encoder} variant, the projected text embedding was added to the sparse prompt input tokens. For the \textbf{mask decoder}, we added the projected embedding to the decoder input tokens before passing them into the transformer blocks. In each case, the remaining components followed their default design: when injecting into the prompt encoder, both the image encoder and mask decoder were unmodified (with the image encoder using standard parallel adapters); when injecting into the mask decoder, the image encoder used only regular adapters, and the prompt encoder was vanilla.

As shown in Table~\ref{tab:ablation_text_injection}, injecting text into the image encoder yielded the best performance, achieving 71.38 mIoU on ADE20K \texttt{1\_64}. Injecting into the prompt encoder resulted in slightly lower performance (71.11), and the mask decoder variant also underperformed relative to the image encoder (70.82). These results suggest that the image encoder is the optimal integration point for semantic conditioning, likely because it shapes the visual features at an early stage of the processing pipeline.

\begin{table}[htbp]
\centering
\begin{tabular}{lc}
\toprule
Text Injection Location & ADE20K \texttt{1\_64} mIoU \\
\midrule
Prompt Encoder only & 71.11 \\
Image Encoder (ours) & \textbf{71.38} \\
Mask Decoder & 70.82 \\
\bottomrule
\end{tabular}
\vspace{1mm}
\caption{Design ablation: where to inject CLIP text embeddings. Injecting into the image encoder yields the best performance under low-data settings.}
\label{tab:ablation_text_injection}
\end{table}

\subsubsection{Adapter-Text Placement Ablation}
\label{sec:adapter_ablation}

We investigated where to inject the text-conditioned adapter within each transformer block—specifically, whether to apply the text-guided adapter to the MLP and MHSA branches or only to the MLP. As shown in Table~\ref{tab:ablation_adapter_variant}, using the semantic extension to both branches resulted in a slight performance drop, with 71.25 mIoU on ADE20K \texttt{1\_64}, compared to 71.38 when restricting the text guidance to the MLP-parallel adapter. This suggested that injecting semantics into the attention pathway might interfere with prompt propagation, potentially disrupting SAM’s ability to localize objects based on spatial cues. By confining text conditioning to the MLP path, we preserved SAM’s spatial reasoning while effectively integrating semantic information.

\begin{table}[htbp]
\centering
\begin{tabular}{lc}
\toprule
Text Adapter Location & ADE20K \texttt{1\_64} mIoU \\
\midrule
MLP-only (ours) & \textbf{71.38} \\
MLP + MHSA & 71.25 \\
\bottomrule
\end{tabular}
\vspace{1mm}
\caption{Adapter-text placement ablation. Injecting semantics into the adapter parallel to the MHSA slightly reduces performance compared to injecting into the MLP block adapter, suggesting that preserving spatial pathways is important.}
\label{tab:ablation_adapter_variant}
\end{table}

\subsection{Results on COD10K: A Novel Vision-Language Benchmark}
\label{sec:cod10k_results}

To further validate our approach, we evaluate SAM-PTx on the COD10K~\cite{fan2020COD10K} dataset—a challenging benchmark for segmenting fine-grained and visually subtle objects. COD10K is especially difficult for SAM, and it is one of the main open-source datasets used to benchmark parameter-efficient fine-tuning (PEFT) methods for SAM~\cite{chen2023SamAdptr, song2024SUSAM}. Its complex backgrounds and camouflaged targets make it a meaningful testbed for evaluating segmentation methods under minimal supervision.

While COD10K is widely used in segmentation research, to the best of our knowledge, this is the first work to adapt it for a vision-language setting by incorporating class names as semantic text prompts.

As shown in Table~\ref{tab:cod10k_results}, SAM-PTx achieves a mean absolute error (MAE) of \textbf{0.0206}, outperforming all prior SAM-based approaches. SAM Adapter~\cite{chen2023SamAdptr} reports an MAE of 0.025, and SU-SAM's~\cite{song2024SUSAM} best-performing variant (the mixed adapter) also reports 0.025. The parallel adapter from SU-SAM, which serves as the architectural foundation for SAM-PTx, was originally reported to have an MAE of 0.054.

In our experiments, we observed differences from the reported results in~\cite{song2024SUSAM}, particularly for the mixed adapter variant. The performance of the parallel adapter was more consistent with our replication attempts. To ensure a fair comparison, we retrained the parallel adapter using our training setup—including the same learning rate, number of epochs, and optimizer configuration—and obtained an improved MAE of 0.0213. While this variant approaches the performance of SAM-PTx, SAM-PTx still achieves the best result at 0.0206, highlighting the benefit of incorporating semantic text guidance during fine-tuning.

\begin{table}[htbp]
\centering
\begin{tabular}{lc}
\toprule
Method & MAE $\downarrow$ \\
\midrule
SAM Adapter~\cite{chen2023SamAdptr} & 0.025 \\
SU-SAM (Mixed Adapter)~\cite{song2024SUSAM} & 0.025 \\
SU-SAM (Parallel Adapter)~\cite{song2024SUSAM} & 0.054 \\
\textbf{Ours (Parallel-Text Adapter)} & \textbf{0.021} \\
\bottomrule
\end{tabular}
\vspace{1mm}
\caption{Segmentation performance on COD10K. SAM-PTx achieves the lowest MAE by incorporating semantic text prompts, demonstrating the benefit of vision-language alignment in SAM fine-tuning.}
\label{tab:cod10k_results}
\end{table}

In contrast, SAM-PTx, trained with frozen CLIP-derived class name embeddings and point prompts, not only improves segmentation quality but also demonstrates that semantic text guidance helps SAM disambiguate fine-grained objects in complex scenes better. This highlights both the effectiveness of SAM-PTx and the broader potential of language integration in fine-tuning SAM for new domains.

\subsection{Qualitative Analysis}

\label{sec:qualitative_analysis}

\begin{figure}[t]
    \centering
    \includegraphics[width=\linewidth]{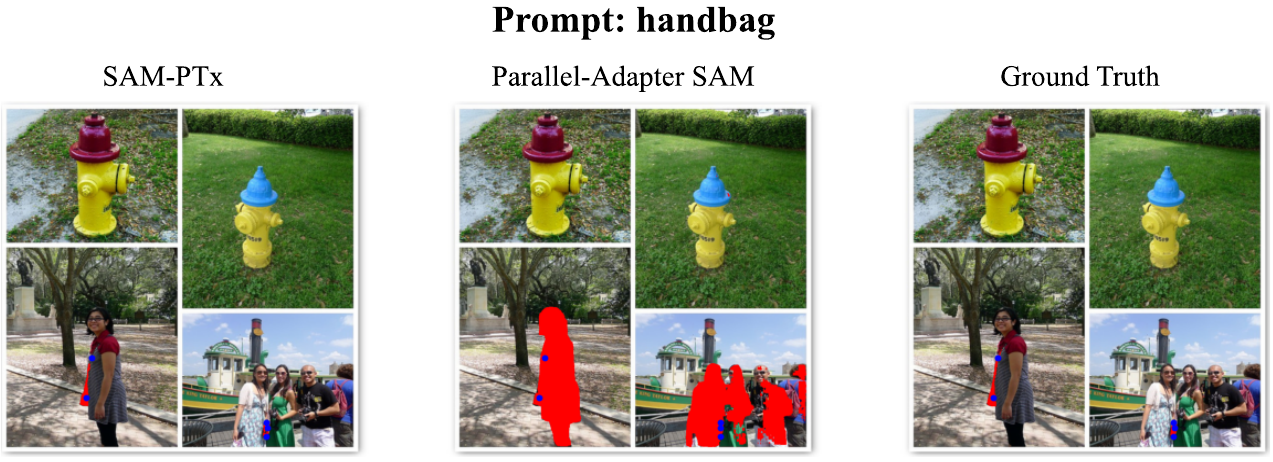}
    \caption{
    \textbf{Category 1 — Edge-case prompts and small objects.}
    The baseline model (middle) often leaks into adjacent areas when point prompts—shown in blue—lie near object edges or thin structures. SAM-PTx (right) maintains tighter object boundaries due to semantic awareness. Ground truth is shown on the left.
    }
    \label{fig:qual_cat1}
\end{figure}

To better understand SAM-PTx's behaviour and the impact of semantic guidance, we analyzed qualitative results across a diverse set of segmentation scenarios. We observed several common patterns, which we categorize below:

\paragraph{Category 1 — Edge-case prompts and small objects.}
In these cases, point prompts lie on the object's boundary or fall near ambiguous or thin regions (e.g., elongated limbs or small objects). These placements can lead to mask leakage into neighbouring areas. The Parallel Adapter variant often spills into adjacent objects or background regions. In contrast, SAM-PTx is better at constraining the mask to the correct object, demonstrating improved robustness to imprecise spatial prompts.
(See Figure~\ref{fig:qual_cat1})

\begin{figure}[t]
    \centering
    \includegraphics[width=\linewidth]{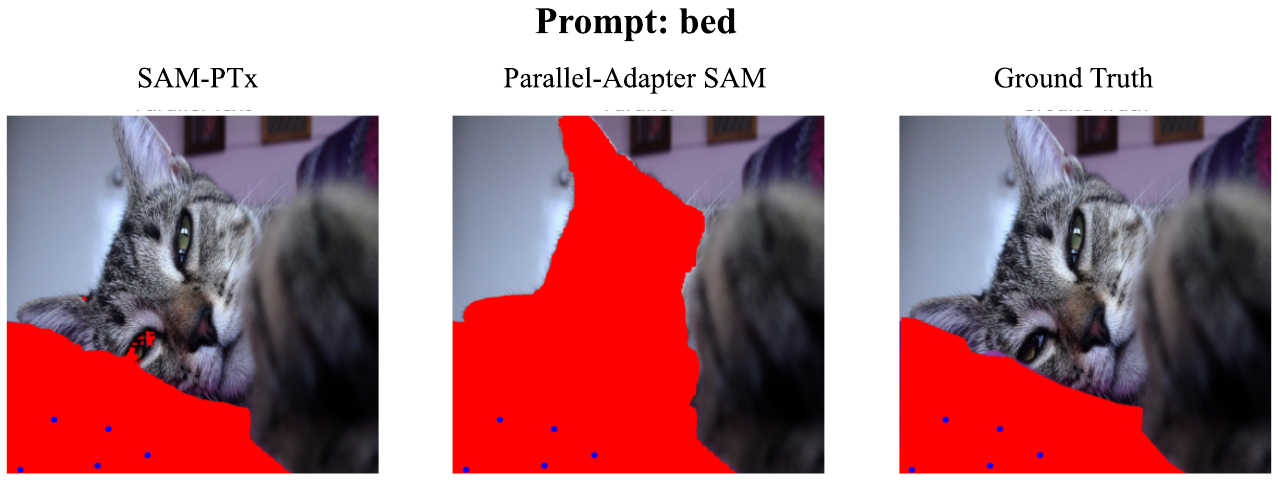}
    \caption{
    \textbf{Category 2 — Good prompts, poor segmentation.}
    Despite well-placed point prompts—shown in blue—the baseline fails to segment the object fully. With class-level semantic conditioning, SAM-PTx produces more complete masks with improved boundary precision.
    }
    \label{fig:qual_cat2}
\end{figure}

\paragraph{Category 2 — Good prompts, poor segmentation.}
All prompts in this category lie cleanly within the object’s interior, yet the segmentation is unsatisfactory. This often results from the model’s inability to accurately recognize the object class or delineate boundaries. Even though text guidance may not fully resolve this in every case, the Parallel-Text Adapter often produces tighter, more complete masks than the baseline.
(See Figure~\ref{fig:qual_cat2})

\begin{figure}[t]
    \centering
    \includegraphics[width=\linewidth]{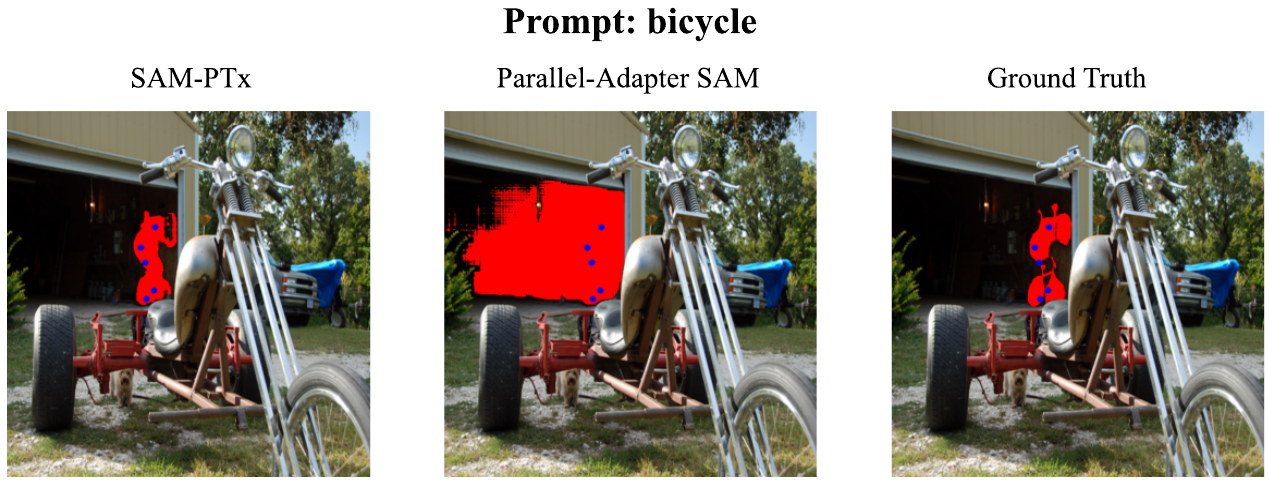}
    \caption{
    \textbf{Category 3 — Mixed prompt placement and boundary spill.}
    Some point prompts—shown in blue—fall near boundaries while others are central. The baseline model produces a mask that spills outside the target object. In contrast, SAM-PTx produces more precise masks that better capture the true shape of the object.
    }
    \label{fig:qual_cat3}
\end{figure}

\paragraph{Category 3 — Mixed prompt placement and boundary spill.}
This category includes cases where some prompts are well-placed while others fall near object edges. Despite sufficient point cues, the baseline model still produces imprecise masks that extend beyond object boundaries. Semantic information helps the model resolve ambiguity in such scenarios, producing more localized masks.
(See Figure~\ref{fig:qual_cat3})

\begin{figure}[t]
    \centering
    \includegraphics[width=\linewidth]{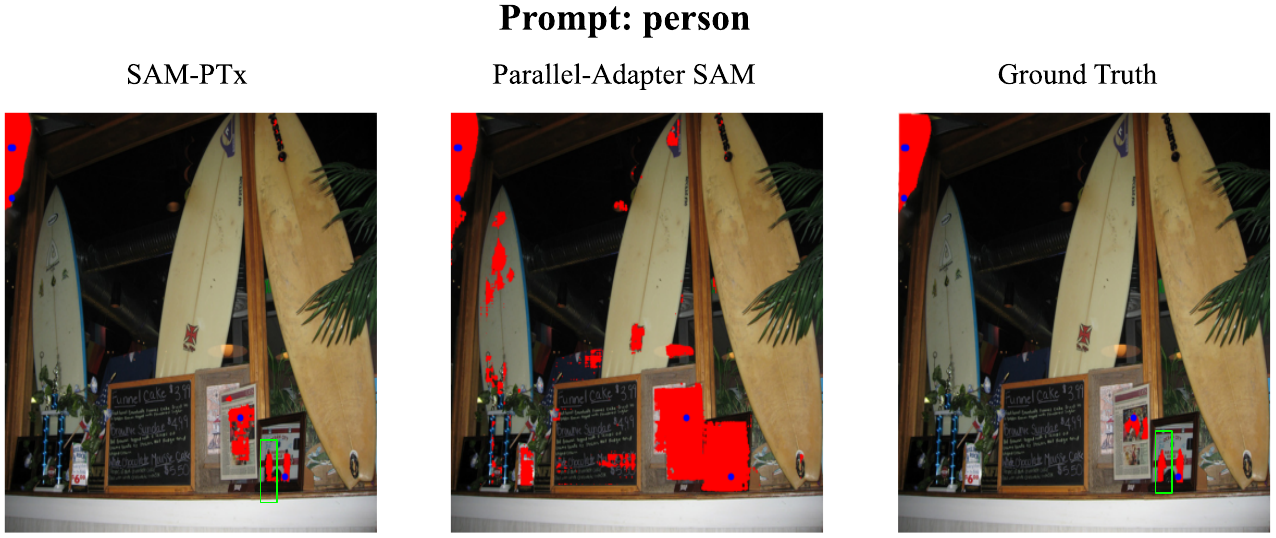}
    \caption{
    \textbf{Category 4 — Unprompted instances.}
    Only a subset of class instances have point prompts (shown in blue). SAM-PTx better segments the additional matching instances—one of which is highlighted in a green box—while the baseline limits itself to prompted regions. Semantic guidance helps in better recognizing class-level repetition.
    }
    \label{fig:qual_cat4}
\end{figure}

\paragraph{Category 4 — Missing prompts for some instances.}
In scenes with multiple instances of a target class, sometimes only a subset of the objects receive point prompts. A strong model should still segment the remaining objects if they match the class name. The Parallel Adapter struggles to identify these unprompted instances, while the Parallel-Text Adapter segments them successfully, leveraging semantic understanding from the class-level text input.
(See Figure~\ref{fig:qual_cat4})

\begin{figure}[t]
    \centering
    \includegraphics[width=\linewidth]{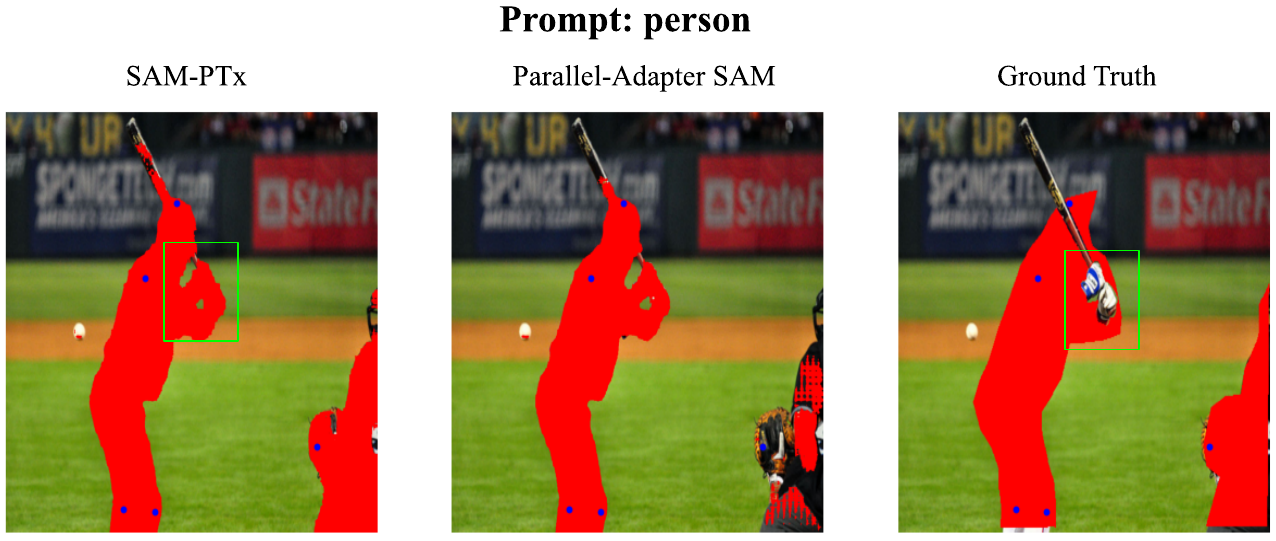}
    \caption{
    \textbf{Category 5 — SAM’s high boundary precision.}
    SAM-PTx can sometimes produce cleaner or more precise segmentations than the ground truth (left), particularly in thin structures or detailed textures (green box). This highlights the strong visual priors of SAM, enhanced by semantic cues.
    }
    \label{fig:qual_cat5}
\end{figure}

\paragraph{Category 5 — SAM’s strong boundary precision.}
Interestingly, we occasionally observe outputs from both SAM variants—especially with text conditioning—that appear cleaner and more precise than the ground truth masks, particularly around fine structures like animal fur or overlapping object edges. This highlights SAM’s inherent ability to capture high-resolution details and suggests that, in some cases, the model may outperform noisy or coarsely labelled human annotations.
(See Figure~\ref{fig:qual_cat5})

\section{Conclusion}
\label{sec:conclusion}

We proposed SAM-PTx, a parameter-efficient framework for incorporating class-level semantic guidance into SAM by injecting frozen CLIP-derived text embeddings into SAM’s image encoder through lightweight adapters. Our design builds on the Parallel Adapter structure from SU-SAM, modifying only the MLP-parallel branch to preserve spatial reasoning while enabling vision-language alignment.

Through experiments on low-data splits of ADE20K, COCO, and the challenging COD10K benchmark, we demonstrated that our Parallel-Text Adapter consistently improves segmentation performance over purely spatial prompt-based baselines. Design ablations further showed that injecting semantic information into the image encoder and confining it to the MLP pathway yields the most effective integration.

Qualitative results revealed that text guidance improves robustness to imprecise prompts, enhances generalization to unprompted object instances, and produces cleaner segmentation boundaries that, in some cases, surpass the quality of human-labelled ground truth. These findings suggest that integrating text semantics into SAM and fine-tuning SAM with them offers a practical and scalable path toward semantically guided segmentation.

Our method introduces minimal architectural changes and requires training only a small fraction of parameters, making it practical for real-world fine-tuning. In future work, we plan to explore joint training with CLIP, more expressive text prompts, and extensions to open-vocabulary and few-shot segmentation. Our findings suggest that SAM-PTx provides a practical and scalable path for integrating semantic guidance and text-based inputs into SAM's architecture and fine-tuning process.
{\small
\bibliographystyle{IEEEtran}
\bibliography{main}
}

\end{document}